\documentclass{article}
\usepackage[utf8x]{inputenc}
\usepackage{user_commands}
\usepackage{amssymb,amsmath}
\usepackage{epsfig, color}
\usepackage{multirow}
\usepackage{algorithm}
\usepackage{algorithmic}
\usepackage{ifthen}
\usepackage{cite}

\newboolean{draftpaper}

\title{Artificial Neural Networks, Symmetries and Differential Evolution}
\author{Onay Urfalioglu, Orhan Arikan\\Department of Electrical and Electronics Engineering, Bilkent University}

\begin{document}
\maketitle


\begin{abstract}
Neuroevolution is an active and growing research field, especially in times of increasingly parallel computing architectures. Learning 
methods for Artificial Neural Networks (ANN) can be divided into two groups. Neuroevolution is mainly based on Monte-Carlo techniques and 
belongs to the group of global search methods, whereas other methods such as backpropagation belong to the group of local search methods. ANN's 
comprise important symmetry properties, which can influence Monte-Carlo methods. On the other hand, local search methods are generally 
unaffected by these symmetries.
In the literature, dealing with the symmetries is generally reported as being not effective or even yielding inferior results.
In this paper, we introduce the so called \emph{Minimum Global Optimum Distance} principle derived from theoretical considerations 
for effective symmetry breaking, applied to offline supervised learning. Using Differential Evolution (DE), which is a popular and robust evolutionary global optimization method, we experimentally show significant global search efficiency 
improvements by symmetry breaking.
\end{abstract}

\section{Introduction}
Artificial Neural Networks (ANN) are general function approximators~\cite{haykin-1998} and can be used to find a functional representation of a data set. Another point of view is that ANN's represent a way of data compression~\cite{Arbib2002}. The compression ratio depends on the number of neurons used in the ANN which encodes the data: the less neurons at the same representation quality, the better the compression. 

The estimation of the ANN-parameters is generally a computationally demanding task~\cite{sirma2001}. The corresponding Maximum-Likelihood derived cost function comprises many local optima. Therefore, local search techniques to find an optimal solution generally fail and only a suboptimal solution is found, which is a local optimum~\cite{haykin-1998}. In addition, local search techniques are mainly sequential methods and parallel implementations are limited. On the other hand, global optimization techniques based on Monte-Carlo methods such as the Genetic Algorithm (GA)~\cite{Goldberg1989,Michalewicz1994}, Covariance Matrix Adaptation Evolution Strategies (CMA-ES)~\cite{Hansen:1996,Hansen:2003} or Differential Evolution (DE)~\cite{Storn95a,Pri96,Vesterstrom2004} are generally very well parallelizable. Differential Evolution is one of the most popular and robust Monte-Carlo global search methods, which outperforms many other evolutionary algorithms on a wide range of problems~\cite{BersiniDLSG96,tusar2007,Xu2007}. DE is successfully used in many engineering problems such as multiprocessor synthesis~\cite{Rae1998}, optimization of radio network designs~\cite{Mendes2006}, training RBF networks~\cite{Liu2005}, training multi layer neural networks~\cite{Ilonen2003} and many others. 

Due to inherent symmetries in the parametric representation of ANN's, there are also multiple \emph{global} optima. The multiple global optima result from point symmetries and permutation symmetries~\cite{Thierens96,Sussmann1992}. The effect of these symmetries on Genetic Algorithms is reported to be minimal and negligable~\cite{Haflidason2009}.
However, there are no reports on the impact of the ANN-symmetries regarding the DE method.
In this paper, we show that DE is very sensitive to multiple global optima. We derive a symmetry breaking operator based on theoretical considerations, which is optimal according to a \emph{Minimum Global Optimum Distance} condition. In experimental studies on offline supervised learning problems, a significant improvement of up to two orders of magnitude is achieved by symmetry breaking in terms of global convergence speed. Comparisons to CMA-ES, which is a state-of-the-art evolutionary method for ANN-learning~\cite{EANT2-CMAES-APP-SiebBoet09,Gomez2008}, show that CMA-ES is outperformed on complex learning problems using smaller networks which represent better compression.

\section{Brief Review of Artificial Feedforward Neural Networks}
We deal with Artificial (Feedforward) Neural Networks (ANN) for approximation of functions $f: [-1,1]^d \rightarrow [-1,1]^q$, having $L$ layers (one input layer, $L-2$ hidden layers and one output layer) and $N_l$ sigmoid type neurons per hidden layer $l$. 
For each neuron $(l,n)$, we denote a parameter vector by 
\begin{equation}
\bm{\eta}^l_{n}=(\bm{w}^l_{n},\tau^l_{n}),
\end{equation}  
where $\bm{w}^l_{n}$ is the weight vector and $\tau^l_{n}$ is the shift scalar. The output of a tanh-type sigmoid neuron $(l,n)$ is calculated by 
\begin{equation}\label{eq:tanh}
 x^l_{n}=\tanh({\bm{w}^l_{n}}^{\top}\bm{x}^{l-1}+\tau^l_{n}),
\end{equation} 
where $\bm{x}^l=(x^l_{1}, \ldots, x^l_{N_l})$ is the output vector of layer $l$. After all hidden layers $l=2,3,...,L-1$ are evaluated, the final output component $\hat{y}_n$ of the output vector $\hat{\bm{y}}$ is calculated by
\begin{equation}\label{eq:lin-ouput-layer}
 \hat{y}_n={\bm{w}^L_{n}}^{\top}\bm{x}^{L-1},\ n=1,..,q.
\end{equation} 
We denote the parameter vector of all neurons in a layer $l$ by $\bm{\lambda}^l$, where
\begin{equation}
 \bm{\lambda}^l=(\bm{\eta}^l_{1},\ldots,\bm{\eta}^l_{N_l}).
\end{equation} 
The total parameter vector of the whole network is given by 
\begin{equation}
\bm{\theta}_a=(\bm{\lambda}^2,\ldots,\bm{\lambda}^{L-1},\bm{w}^L_{1},\ldots,\bm{w}^L_{q}),
\end{equation} 
where $\bm{w}^L=(w^L_{1},\ldots,w^L_{N_L})$ is the vector of the output layer weights. The function defined by the network is denoted by
\begin{equation}
 \hat{\bm{y}}=\Omega(\bm{\theta}_a;\bm{x}),
\end{equation} 
where $\bm{x}$ is the input vector, which equals to the output of the input layer, so that $\bm{x}^1\equiv\bm{x}$.
\\
Assuming additive normal i.i.d. noise on the available data $(\bm{x}_k,\bm{y}_k), k=1,...,K$, the ML-estimate $\hat{\bm{\theta}}_a$ of the parameters $\bm{\theta}_a$ is determined by the least squares solution:
\begin{equation}\label{eq:ANN-argmin}
 \hat{\bm{\theta}}_a=\arg\min_{\bm{\theta}_a}\sum_{k=1}^K(\bm{y}_k-\Omega(\bm{\theta}_a;\bm{x}_k))^{\top}(\bm{y}_k-\Omega(\bm{\theta}_a;\bm{x}_k)).
\end{equation} 
Since the output layer is linear as shown in Eqn.~(\ref{eq:lin-ouput-layer}), the corresponding weights $\bm{w}_{M,n}$ can be determined by a least squares method, as described in~\cite{Masters93}, which we adopt in this paper. This has the advantage that global search is applied only to the non-linear part of the parameter space, which generally speeds up convergence. Consequently, the parameter vector for global optimization $\bm{\theta}$ consists of
\begin{equation}
 \bm{\theta}=(\bm{\lambda}^2,\ldots,\bm{\lambda}^{L-1}).
\end{equation} 
In the following section, we briefly review the DE method.
\section{Brief Review of Differential Evolution}\label{sec:DE-review}
DE is one of the best general purpose evolutionary global optimization methods available. It has only linear complexity and it is known as an efficient global optimization method for continuous problem spaces. The optimization is based on a population of $N_p$ solution candidates $\bm{\theta}_{i},i\in\{1,...,N_p\}$ where each candidate has a position in the $D$-dimensional search space. Initially, the solution candidates are generated randomly according to a uniform distribution within the provided intervals of the search space. The population improves by generating new positions iteratively for each candidate. For each individual $\bm{\theta}_{i,G}$, a new trial position $\bm{u}_{i,G}$ is determined by 
\begin{eqnarray}
\bm{v}_{i,G}&=&\bm{\theta}_{r_{1},G}+F\cdot(\bm{\theta}_{r_{2},G}-\bm{\theta}_{r_{3},G})\label{eq:diff-term}\\
\bm{u}_{i,G}&=&C\left(\bm{\theta}_{i,G},\ \bm{v}_{i,G}\right)\label{eq:DE-trial},
\end{eqnarray}
where $r_1,r_2,r_3$ are pairwise different randomly chosen integers from the discrete set $\{1,...,N_p\}$ and $F$ is a weighting scalar. The vector $\bm{v}_{i,G}$ is used together with $\bm{\theta}_{i,G}$ in the crossover operation, denoted by $C()$. The crossover operator copies coordinates from both $\bm{\theta}_{i,G}$ and~$\bm{v}_{i,G}$ in order to create the trial vector $\bm{u}_{i,G}$. $C$ is provided with the probability $C_r$ to copy coordinates from $\bm{\theta}_{i,G}$, whereby coordinates from $\bm{v}_{i,G}$ are copied with a probability of $1-C_r$ to $\bm{u}_{i,G}$. Only if the new candidate $\bm{u}_{i,G}$ proves to have a lower cost then it replaces $\bm{\theta}_{i,G}$, otherwise it is discarded.\\

DE includes an adaptive range scaling for the generation of solution candidates through the difference term in Equation~(\ref{eq:diff-term}). This leads to a global search with large step sizes in the case where the solution candidate vectors are widely spread within the search space due to a relatively large mean difference vector. In the case of a converging population, the mean difference vector becomes relatively small and this enables efficient fine tuning at the final phase of the optimization process. The crossover operator has a complicated role in the dynamics of the population. For example, it produces rotations that are very important when dealing with separable variables. In some cases, it can help to increase the diversity of the population or it can also speed up the convergence, depending on the problem.

\section{Symmetries in ANN's}
\begin{figure}[h]
   \begin{center}
      \scalebox{0.6}{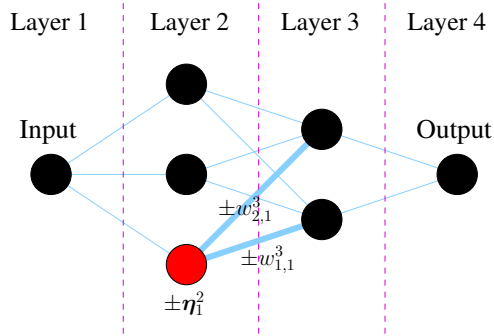}
      \caption{\label{fig:point-symm-example} \it Application of the point symmetry operator $O^2_{1}$, which changes the signs of 
$\bm{\eta}^2_1$-parameters in layer two and $w^3_{n,1}$-parameters in layer three, respectively.}
   \end{center} 
\end{figure}

A \emph{symmetry} is an operator $\Phi$ which applies to the parameter vector $\bm{\theta}$ of the ANN and leaves the 
output of the ANN invariant:
\begin{equation}
 \Omega(\bm{\theta};\bm{x})=\Omega(\Phi(\bm{\theta});\bm{x}) \ \forall \bm{\theta},\bm{x}.
\end{equation} 
Non reducable ANN's comprise two types of symmetries. The first type is a \emph{point symmetry} on the neuron parameter level, since
\begin{equation}
w\tanh(x)=-w\tanh(-x)\ \forall w,x. 
\end{equation}
It follows, that a point symmetry operator $O^l_{n}$ defined by
\begin{equation}\label{eq:point-symm}
 O^l_{n}(\bm{\theta}): \left\{\begin{array}{lcl}
\bm{\eta}^l_{n} & \rightarrow & -\bm{\eta}^l_{n}\\
w^{l+1}_{i,n} & \rightarrow & -w^{l+1}_{i,n},\ i=1,\ldots,N_{l+1}                          
\end{array}\right.
\end{equation} 
applied to the parameters of neuron $(l,n)$ and the $n$-th weight component $w^{l+1}_{i,n}$ to all neurons in the following 
layer $l+1$ does not change the output of the ANN. In Fig.~\ref{fig:point-symm-example}, an example for the application of $O^2_{1}$ 
is given. For each layer, the point symmetry yields $2^{N_l}$ symmetric equivalents of the parameter vector $\bm{\theta}$ due to the point symmetries.

\begin{figure}[h]
   \begin{center}
      \scalebox{0.6}{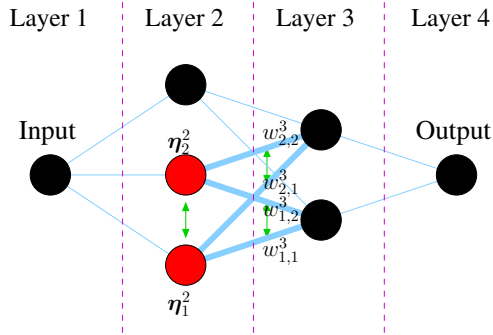}
      \caption{\label{fig:perm-symm-example} \it Application of the permutation symmetry operator $P^2_{1,2}=P^2_{2,1}$, which exchanges 
the parameters $\bm{\eta}^2_1\leftrightarrow\bm{\eta}^2_2$ in layer two and the parameters $w^3_{1,1}\leftrightarrow w^3_{1,2}$, $w^3_{2,1}\leftrightarrow w^3_{2,2}$ in layer three.}
   \end{center} 
\end{figure}

\begin{figure}[h]
   \begin{center}
      \scalebox{0.37}{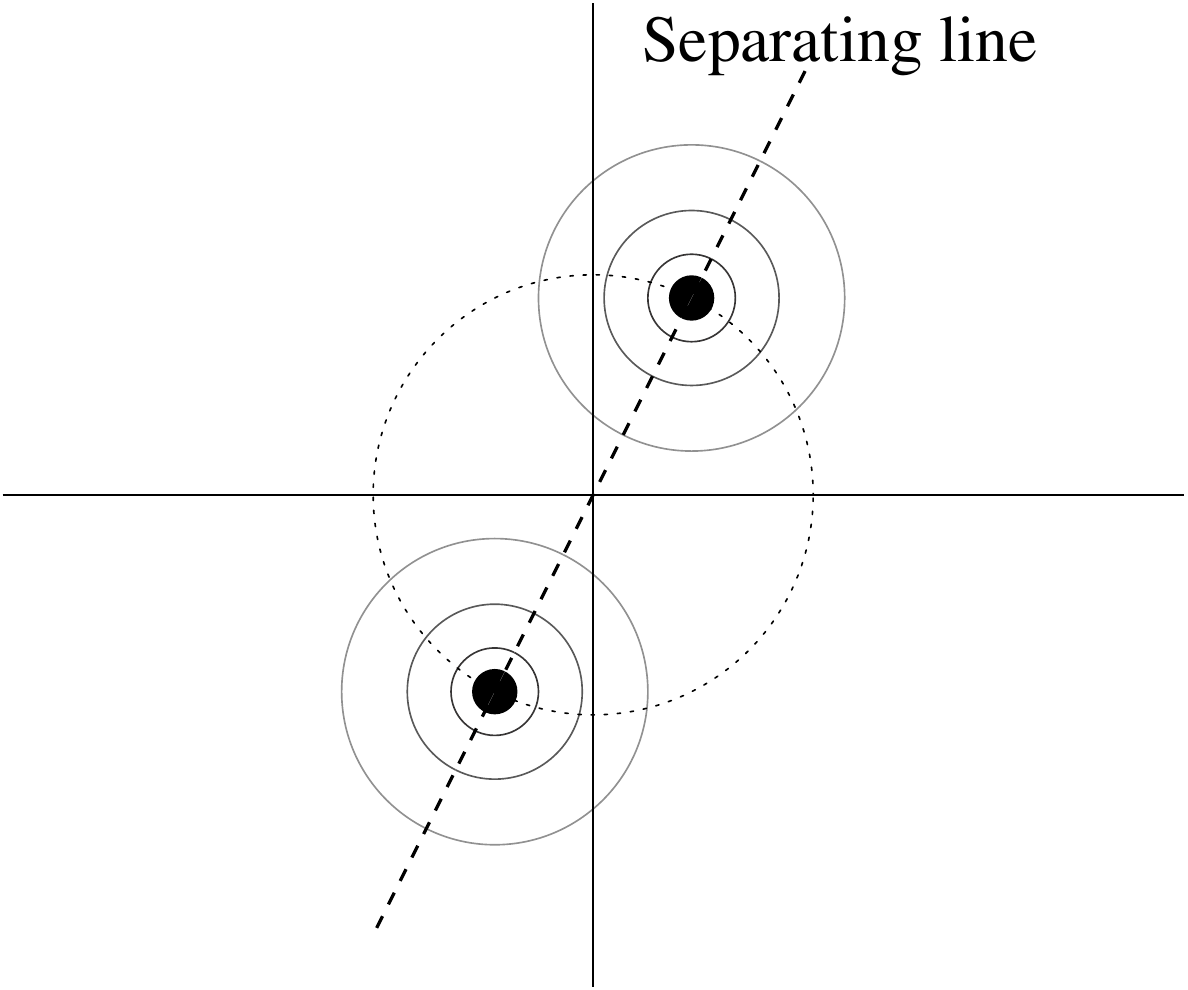}
      \scalebox{0.37}{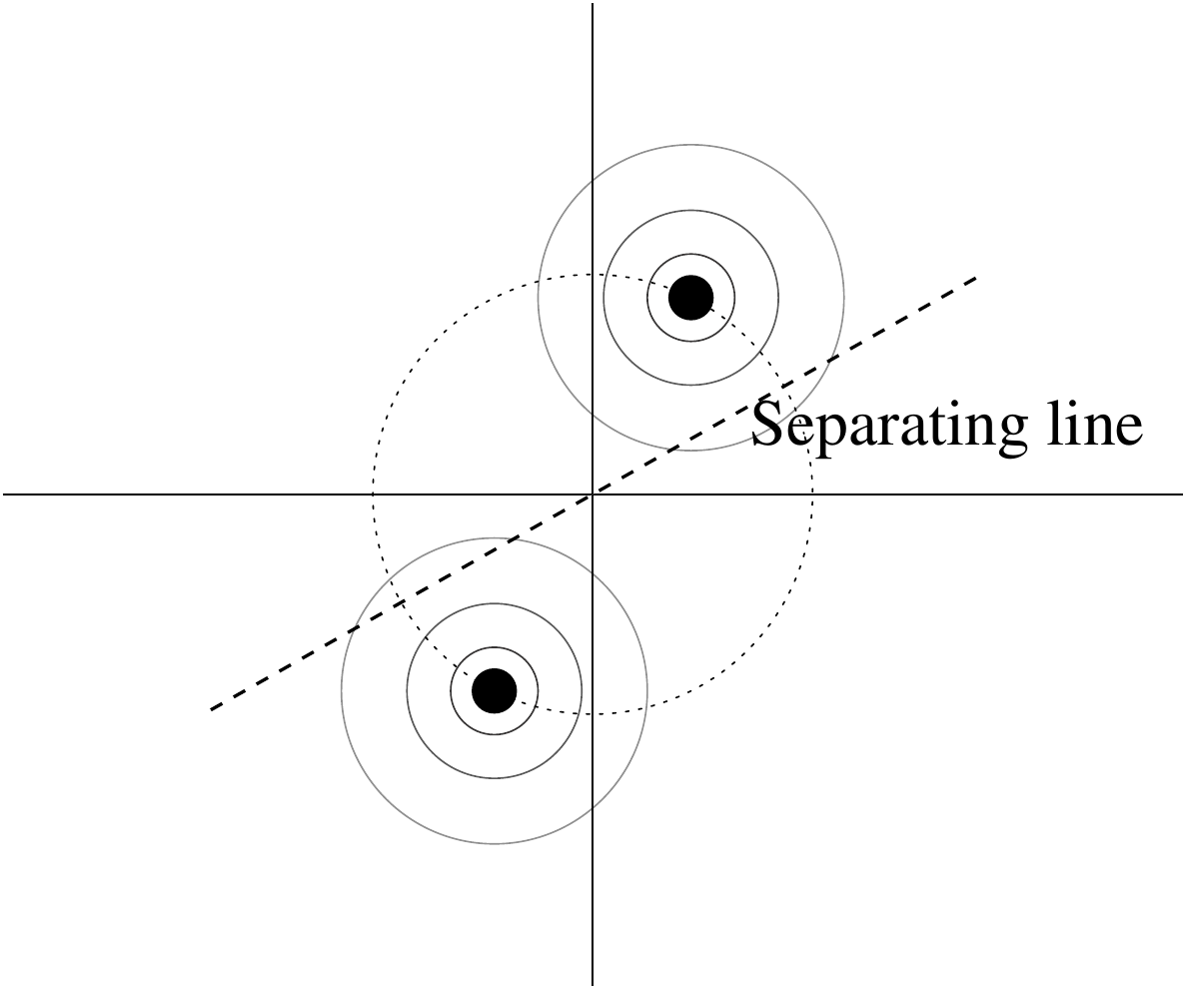}
      \scalebox{0.37}{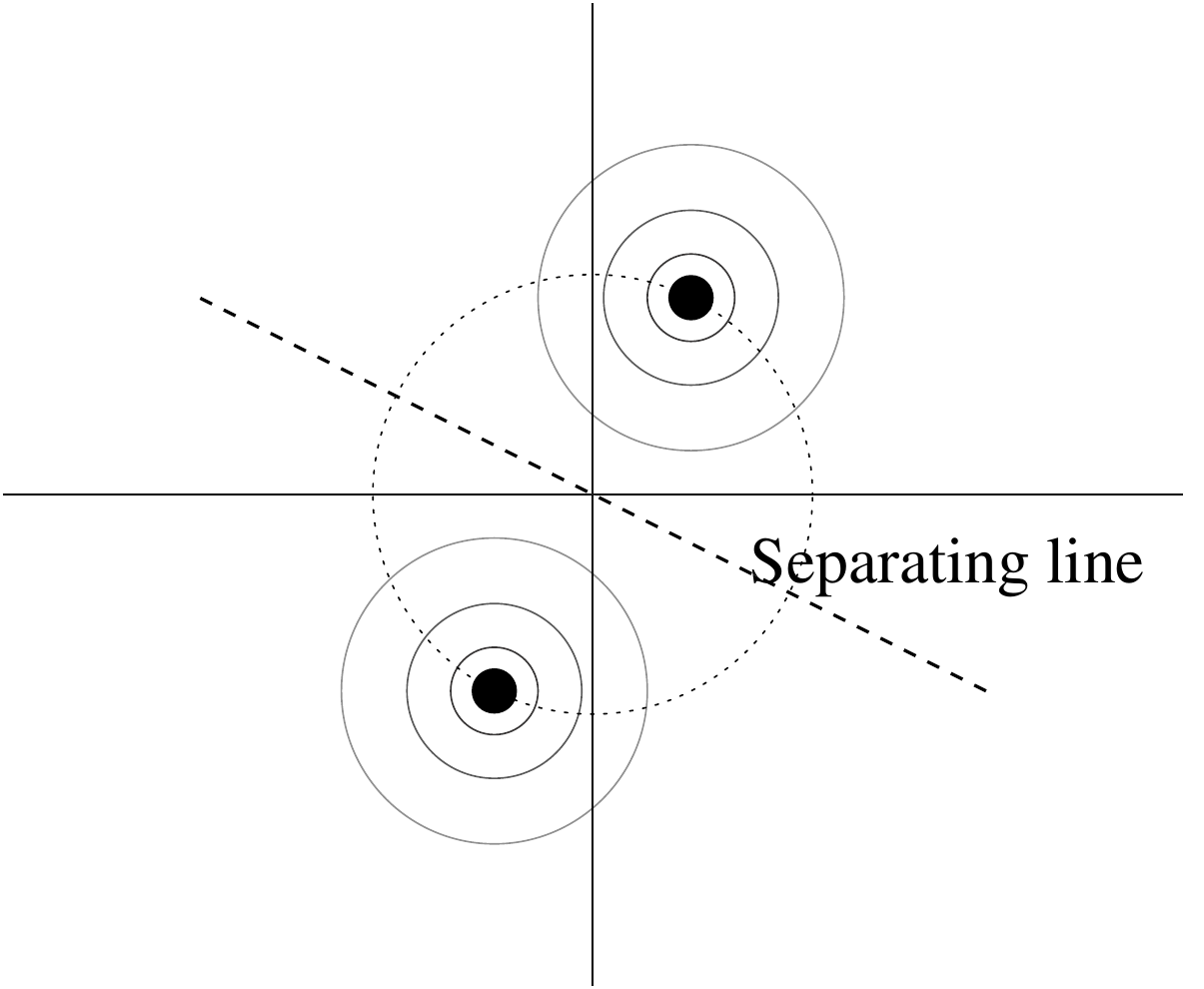}
      \caption{\label{fig:point-symm-go1} \it Example for a point symmetry in 2-D, where $f(\bm{\theta})=f(-\bm{\theta})\ \forall \bm{\theta}$. The plots show worst case (left), suboptimal (middle) and optimal separation lines (right) for point symmetry breaking. The separating line divides the parameter space in two parts, where each partition contains a global optimum ($\bar{\bm{\theta}}$ and $-\bar{\bm{\theta}}$).}
   \end{center} 
\end{figure}

The second type of symmetry is a \emph{permutation symmetry} by the neuron parameters $\bm{\eta}$ and corresponding weight 
parameters in the next layer. A permutation operator $P^l_{j,k}$ defined by 
\begin{equation}\label{eq:perm-symm}
 P^l_{j,k}(\bm{\theta}):\left\{\begin{array}{lcl}
\bm{\eta}^l_{j}\leftrightarrow\bm{\eta}^l_{k}\\
w^{l+1}_{i,j}\leftrightarrow w^{l+1}_{i,k},\ i=1,\ldots,N_{l+1}
\end{array}\right.
\end{equation} 
leaves the output invariant. Note that $P^l_{j,k}=P^l_{k,j}$. In Fig.~\ref{fig:perm-symm-example}, the application of $P^2_{1,2}=P^2_{2,1}$ is illustrated. In each layer, 
there are $N_l!$ symmetric equivalents of the parameter vector $\bm{\theta}$ due to the permutation symmetries. The total count of symmetric equivalents per layer $l$ is $2^{N_l}N_l!$.
Another important property is that the length of the vector $\bm{\theta}$ is invariant under such symmetry operators:
\begin{equation}\label{eq:length-invar}
||\Phi(\bm{\theta})||=||\bm{\theta}||\ \forall \Phi.
\end{equation}  
As a result, all symmetric equivalents of a global optimum lie on a hypersphere.
Since such symmetries multiply the local and global optima count in the parameter space, the ultimate goal of symmetry breaking is to reduce the total number of local optima in the parameter space by avoiding all but one symmetrically equivalent space partitions. In this case,
there are infinitely many ways for symmetry breaking by using the operators $O^l_{n}$ and $P^l_{j,k}$. The differences arise 
from the \emph{condition} on which these operators are to be applied. Fig.~\ref{fig:point-symm-go1} 
shows different ways for 
breaking a point symmetry in relation to the global optimum. It can be seen that the symmetry invariant region which has 
maximum distances to the global optima enables the optimal separation or partitioning. This way, an optimal isolation between all 
symmetric equivalents of the global optima is achieved. As a result, the influence of other neighbouring global optima is decreased to a minimum, which maximizes the attraction of the global optimum of the selected partition.
It can be easily shown that such a symmetry invariant region is the result
of the following separation condition
\begin{equation}\label{eq:sep-cond1}
\hat{\bm{\theta}}=\left\{\begin{array}{lcl}
\bm{\theta} & \text{for} & ||\bm{\theta}-\bar{\bm{\theta}}|| \leq ||-\bm{\theta}-\bar{\bm{\theta}}||\\
-\bm{\theta} & \text{for} & ||\bm{\theta}-\bar{\bm{\theta}}|| > ||-\bm{\theta}-\bar{\bm{\theta}}||,
\end{array}\right.
\end{equation} 
where $\bar{\bm{\theta}}$ denotes the selected global optimum, $\bm{\theta}$ the parameter vector before 
symmetry breaking and $\hat{\bm{\theta}}$ the parameter vector after symmetry breaking. Taking into account the additional permutation symmetry, which is independent from the point symmetry, the operator $\Phi$ is generally composed of a chain of point symmetry and permutation symmetry operators. As an example $\Phi=O^2_{2}\circ P^3_{2,4}$, 
applies a permutation symmetry followed by a point symmetry operator. We denote the set of all possible symmetry operators by $\mathcal{S}$. We generalize the separation condition~(\ref{eq:sep-cond1}) and define the following ideal separation, or, in other words, symmetry breaking
\begin{equation}\label{eq:sep-cond}
\hat{\Phi}=\arg\min_{\Phi\in \mathcal{S}}||\Phi(\bm{\theta})-\bar{\bm{\theta}}||,\ \hat{\bm{\theta}}=\hat{\Phi}(\bm{\theta}).
\end{equation} 
This means, the ideal separation selects the symmetry operator $\hat{\Phi}$ from the set of all possible symmetry operators which minimizes the distance of the parameter vector $\hat{\bm{\theta}}$ to a global optimum $\bar{\bm{\theta}}$.
\subsection{Approximation of the ideal separation}
There are two problems for the practical application of the ideal separation. First, the global optimum is not known a priori. However, iterative algorithms like DE produce intermediate results at each iteration, which can be regarded as an approximation of the global optimum. This approximation becomes better with increasing iteration number. Therefore, we propose to choose the best individual of the DE-population $\tilde{\bm{\theta}}$ at each iteration as an approximation for the global optimum. Second, the brute force method for finding an optimal solution to~(\ref{eq:sep-cond}) has exponential complexity. Instead of trying to find the optimal symmetry breaking operator, we propose to apply only one single symmetry operator $O^l_{n}$ or $P^l_{j,k}$ at a time. The symmetry operator, layer and neuron are randomly selected. The proposed heuristic only approximates $\hat{\Phi}$, but this approximation improves with the iteration count. For each neuron $(l,n)$, we define a symmetry relevant parameter block $\bm{\beta}^l_n$ as
\begin{equation}
 \bm{\beta}^l_n=(\bm{\eta}^l_{n},w^{l+1}_{1,n},\ldots,w^{l+1}_{N_{l+1},n}), \ l=2,...,L-2\ \text{and }\bm{\beta}^{L-1}_n=(\bm{\eta}^{L-1}_{n}).
\end{equation}
Given a parameter vector $\bm{\theta}$ and an approximation of the global optimum $\tilde{\bm{\theta}}$ with corresponding parameter blocks $\tilde{\bm{\beta}}^l_n$, the pseudocode~(\ref{alg:heuristic}) describes the proposed heuristic.
\begin{algorithm}[h]
{\small
\caption{\it Proposed heuristic for breaking symmetry. Parameters $\mu,l,n,m$ are sampled from corresponding discrete uniform distributions $U(.)$. A symmetry operator $X$ is only applied to the parameter vector $\bm{\theta}$ when it decreases the distance to the global optimum $\tilde{\bm{\theta}}$, i.e., $||X(\bm{\theta})-\tilde{\bm{\theta}}||<||\bm{\theta}-\tilde{\bm{\theta}}||$. Algorithm input: $\bm{\theta}$ and $\tilde{\bm{\theta}}$, effect: (eventually) modify parameter vector $\bm{\theta}$.}\label{alg:heuristic}
\begin{algorithmic}
\STATE {{\bf symmetry operator selection}: sample $\mu\sim U(\{0,1\})$}
\STATE {{\bf hidden layer selection}: sample $l=U(\{2,...,L-1\})$}
\STATE {{\bf neuron selection}: sample $n=U(\{1,\ldots,N_l\})$ in hidden layer $l$}
\IF {$\mu=0$}
    \STATE {{\bf [point symmetry operator selected]}}
    \STATE {// would the point symmetry operator $O^l_{n}$ decrease the distance? ($||O^l_{n}(\bm{\theta})-\tilde{\bm{\theta}}|| \stackrel{?}{<} ||\bm{\theta}-\tilde{\bm{\theta}}||$)}
    \STATE calculate distance-square for NOT applying the operator $O^l_{n}$: $D_1=||\bm{\beta}^l_n-\tilde{\bm{\beta}}^l_n||^2$
    \STATE calculate distance-square for applying the operator $O^l_{n}$: $D_2=||-\bm{\beta}^l_n-\tilde{\bm{\beta}}^l_n||^2$
    \IF {$D_1 > D_2$} 
        \STATE apply point symmetry operator $O^l_{n}$: set $\bm{\beta}^l_n=-\bm{\beta}^l_n$
    \ENDIF
\ELSE 
    \STATE {{\bf [permutation symmetry operator selected]}}
    \STATE {{\bf second neuron selection}: sample $m=U(\{1,\ldots,N_l\})$ in hidden layer $l$}
    \STATE {// would the permutation operator $P^l_{m,n}$ decrease the distance? ($||P^l_{m,n}(\bm{\theta})-\tilde{\bm{\theta}}|| \stackrel{?}{<} ||\bm{\theta}-\tilde{\bm{\theta}}||$)}
    \STATE calculate distance-square for NOT applying the operator $P^l_{m,n}$: $D_1=||\bm{\beta}^l_n-\tilde{\bm{\beta}}^l_n||^2+||\bm{\beta}^l_m-\tilde{\bm{\beta}}^l_m||^2$
    \STATE calculate distance-square for applying the operator $P^l_{m,n}$: $D_2=||\bm{\beta}^l_n-\tilde{\bm{\beta}}^l_m||^2+||\bm{\beta}^l_m-\tilde{\bm{\beta}}^l_n||^2$
    \IF {$D_1 > D_2$} 
        \STATE apply permutation symmetry operator $P^l_{m,n}$: swap $\bm{\beta}^l_m \leftrightarrow \bm{\beta}^l_n$
    \ENDIF
\ENDIF
\end{algorithmic}}
\end{algorithm}
A property of this heuristic is that the symmetry is not completely and not uniquely broken and the resulting modified parameters may belong to different partitions over time. As also $\tilde{\bm{\theta}}$ changes over time, the final convergence result will be a random global optimum among all other possible global optima.

The proposed symmetry breaking method is always applied on each individual's position $\bm{\theta}_{i,G}$ (see Eqn.~(\ref{eq:DE-trial})) at each iteration prior to the application of Eqn.~(\ref{eq:DE-trial}).
\section{Experiments}\label{sec:experiments}
In this section, we introduce results of experiments to demonstrate the performance improvements by symmetry breaking. The following methods are compared in regression and classification tests: Differential Evolution (DE), Covariance Matrix Adaptation Evolution Strategies (CMA-ES) and DE with symmetry breaking (DE-SB).
With a $D$-dimensional parameter space, all tests are performed with following settings:
\begin{itemize}
 \item DE, DE-SB settings: $F=0.5$, $C_r=0.9$, initial population is randomly generated in $D$-dim. hypercube $[-1,1]^D$ (uniformly),
 \item CMA-ES settings: we used suggested settings for enhanced global search abilities, mentioned in the C-code reference implementation.
\end{itemize}
Given a parameter $\bm{\theta}$ and a data set $(\bm{x}_i,\bm{y}_i)$, the ANN produces a cost $\epsilon$ defined by the Mean Squared Error (MSE), which is derived from Eqn.~(\ref{eq:ANN-argmin}):
\begin{equation}\label{eq:costf}
 \epsilon=\frac{1}{K\cdot q}\sum_{k=1}^K(\bm{y}_k-\Omega(\bm{\theta};\bm{x}_k))^{\top}(\bm{y}_k-\Omega(\bm{\theta};\bm{x}_k)).
\end{equation} 

In order to limit the $D$-dimensional parameter space to a feasible region, we apply a penalty approach. Due to the length-invariance by the symmetry operators as shown in Eqn.~(\ref{eq:length-invar}), the feasible region is defined by a hypersphere. In case of $||\bm{\theta}|| > \sqrt{D}$, the cost function~(\ref{eq:costf}) is evaluated at a rescaled parameter vector $\frac{\bm{\theta}}{||\bm{\theta}||}$ and a penalty term $50(||\bm{\theta}||-\sqrt{D})$ is added to the cost $\epsilon$. 
\subsection{Regression problems}\label{sec:regressionproblems}
\begin{figure*}[htp]
   \begin{center}
      \scalebox{0.33}{\includegraphics{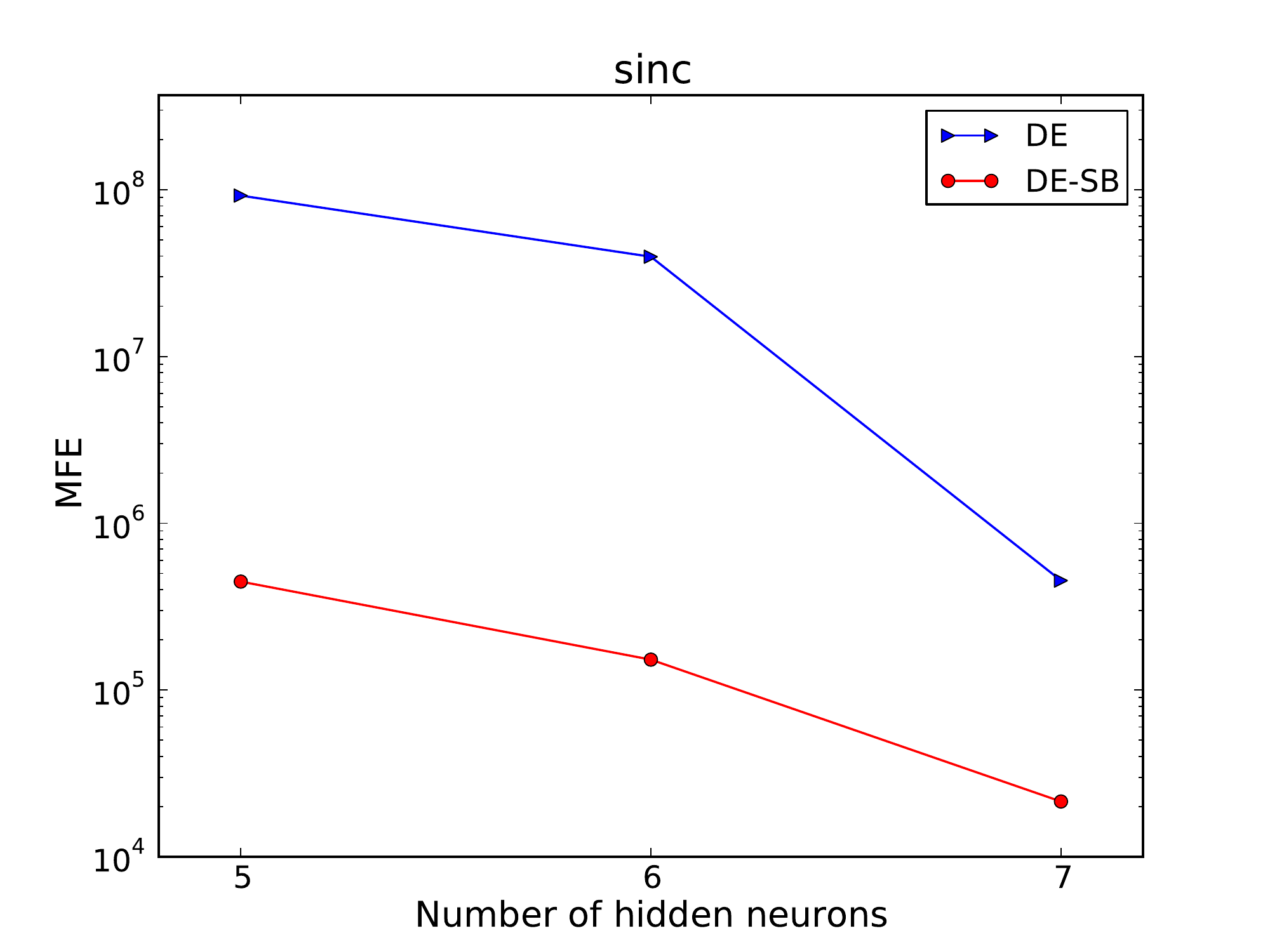}}
      \scalebox{0.33}{\includegraphics{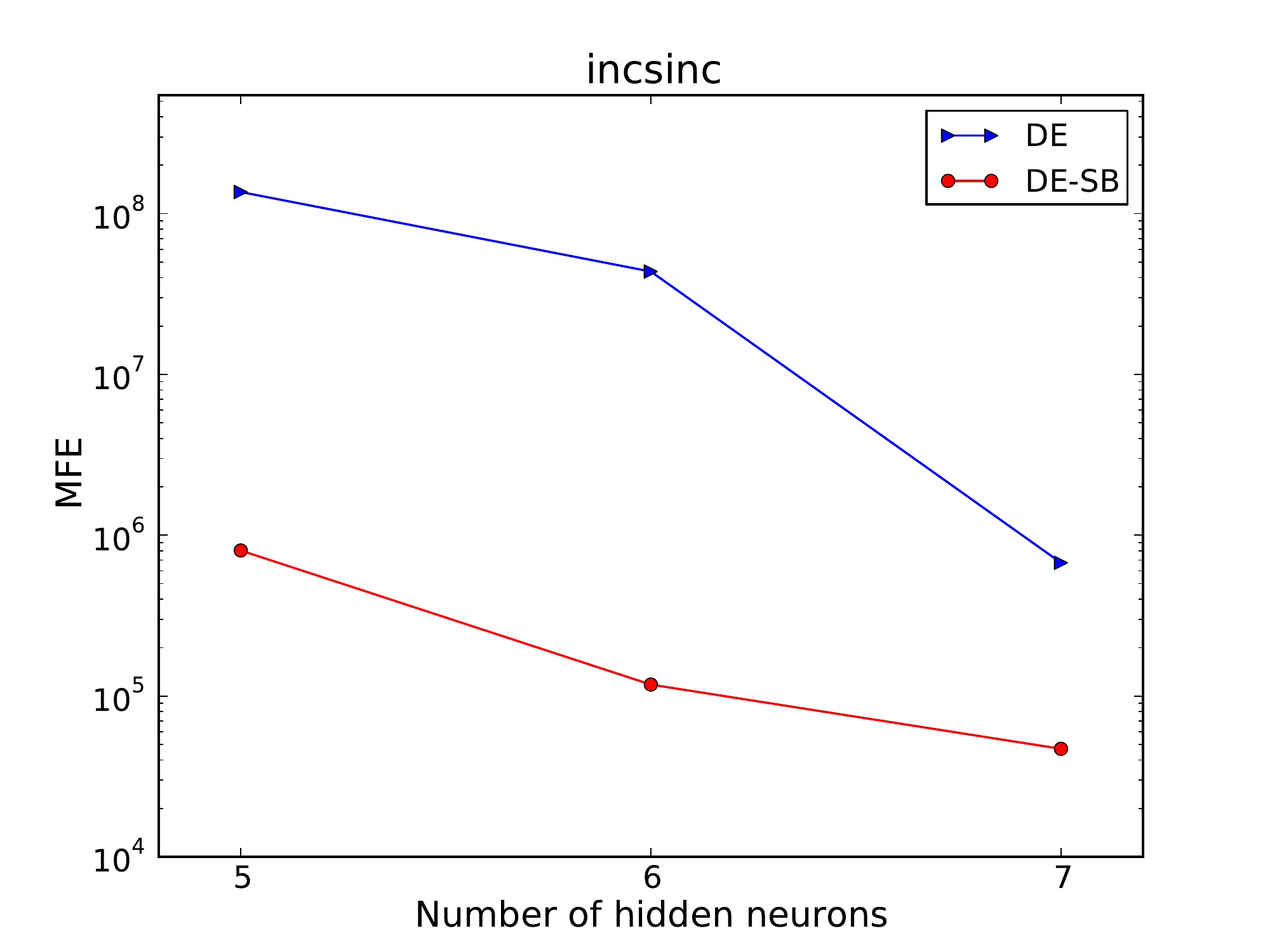}}
      \caption{\label{fig:mfes} \it Required mean function evaluations (MFEs) to find the global optimum 
depending on the number of hidden neurons $N_2$. The corresponding network topology is $(1-N_2-1)$.}
   \end{center}
\end{figure*}

All utilized target functions are defined over $\bm{x}\in[-1,1]^d$ and map to $[-1,1]$. A data set, which consists of training and test samples, is generated by sampling $\bm{x}_i$ from a $d$-dimensional uniform distribution $U^d(-1,1)$
\begin{equation}
 \bm{x}_i\sim U^d(-1,1),
\end{equation} 
 and adding normal distributed noise with zero mean and variance $\sigma^2$ to the function values $y_i$
\begin{equation}
 y_i=f(\bm{x}_i)+\mu, \ \mu\sim \mathcal{N}(0, \sigma^2),\ \sigma=5\!\!\times\!\! 10^{-3}.
\end{equation}
All data sets of regression problems contain 200 training samples and 200 test samples.
Tab.~\ref{tab:regression-funcs} shows the utilized functions and corresponding data sets for regression tests.
\begin{table}[b]
\caption{\label{tab:regression-funcs} \it Target funtions, corresponding data sets, error thresholds and typical test set MSE's of function regression experiments. There was no significant variation of the MSE results by the choice of the learning method.}
\centerline{\tiny
\begin{tabular}{c|c|c|c|c}
 function & data set & training / test samples & $\epsilon_0$ & test set MSE\\
\hline
$(x+0.5)^2(0.1+(x+0.65)^2)$ & {\bf syn5} & 200 / 200 & $5\!\!\times\!\! 10^{-5}$ & $6.27\!\!\times\!\! 10^{-5}\pm 3.2\!\!\times\!\! 10^{-6}$\\
$\sin(10x)/(10x)$ & {\bf sinc} & 200 / 200 & $5\!\!\times\!\! 10^{-5}$ & $5.99\!\!\times\!\! 10^{-5}\pm 7.6\!\!\times\!\! 10^{-6}$\\
$x/2+\sin(10x)/(10x)$ & {\bf incsinc} & 200 / 200 & $5\!\!\times\!\! 10^{-5}$ & $5.88\!\!\times\!\! 10^{-5}\pm 7.1\!\!\times\!\! 10^{-6}$\\
$\sin(5r)/(15r),\ r=\sqrt{x^2+y^2}$ & {\bf sinc2d} & 200 / 200 & $5\!\!\times\!\! 10^{-5}$ & $5.74\!\!\times\!\! 10^{-5}\pm 6.7\!\!\times\!\! 10^{-6}$\\
\end{tabular}}
\end{table}
The population size $N_p$ depends on the problem and is manually adapted accordingly. For each problem, the global optimization is applied in 50 independent runs and the mean required number of ANN-evaluations (MFE) and corresponding standard deviation $\sigma_{\text{MFE}}$ to reach the error threshold $\epsilon_0$ are determined. The population size manually is adapted such that no one of the 50 runs fails to reach the error threshold and the MFE is kept minimal. We declare a global optimum as found by reaching the error threshold. We define a robustness $\rho$ as 
\begin{equation}
 \rho=\frac{\text{number of successfull runs}}{\text{total number of runs}}.
\end{equation} 
Results for the most complex settings (least number of hidden neurons) are shown in Tab.~\ref{tab:regr-res}. Fig.~\ref{fig:mfes} shows results for the required number of ANN-evaluations (MFE) over the number of hidden neurons used in the ANN. 
\begin{table}[htp]
\caption{\label{tab:regr-res} \it Results of Mean ANN (Function) Evaluations (MFE) from 50 independent runs and population sizes of regression tests. Better values are highlighted in boldface. Note that on {\bf sinc} (1-5-1) and {\bf incsinc} (1-5-1), CMA-ES failed to find the global optimum on some runs, even though a large population size of 10000 was used. On {\bf sinc2d} (2-3-1-3-1), CMA-ES failed in all 50 runs.}
\centerline{\tiny
\begin{tabular}{c|c|c|c|c|c}
         &          &           DE                          &     DE-SB                                & CMA-ES  & \\         
data set & topology & [population size, robustness] MFE & [population size, robustness] MFE & [population size, robustness] MFE & $\frac{\text{MFE (DE)}}{\text{MFE (DE-SB)}}$ \\
\hline
{\bf syn5} & 1-3-1 & $[80,\bm{1}]7.30\!\!\times\!\! 10^4\pm 1.8\!\!\times\!\! 10^4$ & $[80,\bm{1}]3.20\!\!\times\!\! 10^4\pm9.4\!\!\times\!\! 10^3$ & $[160,\bm{1}]\bm{1.17\!\!\times\!\! 10^4\pm2.6\!\!\times\!\! 10^3}$& $2.3$\\
\hline
{\bf sinc} & 1-5-1 & $[1400,\bm{1}]9.22\!\!\times\!\!10^7\pm3.1\!\!\times\!\!10^7$ & $[160,\bm{1}]\bm{4.47\!\!\times\!\!10^5\pm2.8\!\!\times\!\!10^5}$ & ${\color{red}[10^4,0.36]3.78\!\!\times\!\!10^6\pm3.7\!\!\times\!\!10^5}$ & $206$\\
{\bf sinc} & 1-6-1 & $[800,\bm{1}]4.19\!\!\times\!\!10^7\pm2.2\!\!\times\!\!10^7$ & $[60,\bm{1}]1.52\!\!\times\!\!10^5\pm4.5\!\!\times\!\!10^4$ & $[100,\bm{1}]\bm{2.67\!\!\times\!\!10^4\pm5.3\!\!\times\!\!10^3}$ & $276$\\
\hline
{\bf incsinc} & 1-5-1 & $[1600,\bm{1}] 1.36\!\!\times\!\!10^8\pm6.0\!\!\times\!\!10^7$ & $[200,\bm{1}]\bm{8.04\!\!\times\!\!10^5\pm1.8\!\!\times\!\!10^5}$ & ${\color{red}[10^4,0.8] 7.76\!\!\times\!\!10^6\pm1.9\!\!\times\!\!10^6}$ & $169$\\
{\bf incsinc} & 1-6-1 & $[800,\bm{1}] 4.36\!\!\times\!\!10^7\pm2.6\!\!\times\!\!10^7$ & $[56,\bm{1}]1.18\!\!\times\!\!10^5\pm5.0\!\!\times\!\!10^4$ & $[100,\bm{1}]\bm{2.86\!\!\times\!\!10^4\pm8.2\!\!\times\!\!10^3}$ & $370$\\
\hline
{\bf sinc2d} & 2-3-1-3-1 & $[120,\bm{1}]1.93\!\!\times\!\!10^5\pm4.2\!\!\times\!\!10^4$ & $[120,\bm{1}]\bm{1.22\!\!\times\!\!10^5\pm2.4\!\!\times\!\!10^4}$ & {\color{red}[$10^4,0$] NA} & $1.58$
\end{tabular}}
\end{table}

\subsection{Classification problems}\label{sec:classproblems}
The following classification problems are used: {\bf iris} data set~\cite{Fisher1988,dasarathy1980irisnosing} , {\bf tic-tac-toe} data set~\cite{Aha1991}, {\bf balance}~\cite{balance-data-set-klahr1978representation,Hume1994} and a more challanging problem defined by the {\bf two-spirals}~\cite{Lang1988} data set. Samples are divided into a training set and a test set following the format $a/b$. The selection process iteratively and sequentially puts $a$ samples into the training set and the following $b$ samples into the test set, until no samples are available. A winner-takes-all scheme is applied to distinguish different classes. The output vector $\bm{y}$ of a sample designating class $i$ has the following format
\begin{equation}
 y_j=\left\{
\begin{matrix}
1 & \text{for}\ j=i\\
0 & \text{else.}             
\end{matrix}\right.
\end{equation} 
As in the case of function regression experiments, on each classification problem, a predefined error threshold $\epsilon_0$ is used as a termination criterion for the learning process. Tab.~\ref{tab:class-def} shows the error thresholds and typical classification sucess rates. 
\begin{table}[htp]
\caption{\label{tab:class-def} \it Data sets, number of training and test samples, the sample selection format, corresponding error thresholds and typical classification sucess rates of classification experiments.}
\centerline{\tiny
\begin{tabular}{c|c|c|c|c}
data set & training / test samples & sample selection & $\epsilon_0$ & test set classification success (\%)\\
\hline
{\bf iris} & 75/75 & 1/1 & $0.011$ & $99.4\pm 0.6$\\
{\bf tic-tac-toe} & 479/479 & 1/1 & $0.08$ &$92.1 \pm 0.8$\\
{\bf balance} & 312/313 & 1/1 & $0.001$ &$100\pm 0$\\
{\bf two-spirals} & 97/97 & 2/2 & $0.07$ &$91.2 \pm 1.4$\\
\end{tabular}}
\end{table}
Tab.~\ref{tab:class-res} shows the results on the classification data sets. 
\begin{table}[htp]
\caption{\label{tab:class-res} \it Results and population sizes of classification tests. Better values are highlighted in boldface. Note that on {\bf balance}, CMA-ES failed to find the global optimum on some runs, even though a large population size of 10000 was used. On {\bf two-spirals} (2-10-1-10-2) and {\bf two-spirals} (2-9-1-9-2), CMA-ES failed in all 50 runs.}
\centerline{\tiny
\begin{tabular}{c|c|c|c|c|c}
         &          &           DE         &     DE-SB             & CMA-ES  & \\         
data set & topology &[population size, robustness] MFE & [population size, robustness] MFE & [population size, robustness] MFE & $\frac{\text{MFE (DE)}}{\text{MFE (DE-SB)}}$\\
\hline
{\bf iris} & 4-3-3 & $[40,\bm{1}]1.46\!\!\times\!\! 10^4\pm 3.3\!\!\times\!\! 10^3$ & $[40,\bm{1}]1.24\!\!\times\!\! 10^4\pm2.5\!\!\times\!\! 10^3$ & $[20,\bm{1}]\bm{3.74\!\!\times\!\! 10^3\pm 4.2\!\!\times\!\! 10^2}$ & $1.18$\\
\hline
{\bf tic-tac-toe} & 9-8-2 &  $[230,\bm{1}]2.70\!\!\times\!\! 10^6\pm 4.7\!\!\times\!\! 10^5$ & $[160,\bm{1}]9.74\!\!\times\!\! 10^5\pm2.2\!\!\times\!\! 10^4$ & $[800,\bm{1}]\bm{3.29\!\!\times\!\! 10^5\pm 3.9\!\!\times\!\! 10^4}$& $2.78$ \\
\hline
{\bf balance} & 4-5-1-5-3 & $[520,\bm{1}]1.38\!\!\times\!\! 10^7\pm 9.2\!\!\times\!\! 10^6$ & $[160,\bm{1}]\bm{8.06\!\!\times\!\! 10^5\pm3.2\!\!\times\!\! 10^5}$ & {\color{red}[$10^4,0.78]5.67\!\!\times\!\! 10^6\pm 5.6\!\!\times\!\! 10^6$} & $17.1$\\
\hline
{\bf two-spirals} & 2-10-1-10-2 & $[200,\bm{1}]2.24\!\!\times\!\! 10^7\pm 4.8\!\!\times\!\! 10^6$ & $[120,\bm{1}]\bm{6.74\!\!\times\!\! 10^6\pm2.5\!\!\times\!\! 10^6}$ & {\color{red}[$10^4,0]$ NA} & $3.32$\\
{\bf two-spirals} & 2-9-1-9-2 & $[240,\bm{1}]4.26\!\!\times\!\! 10^7\pm 1.5\!\!\times\!\! 10^7$ & $[120,\bm{1}]\bm{8.40\!\!\times\!\! 10^6\pm2.6\!\!\times\!\! 10^6}$ & {\color{red}[$10^4,0]$ NA} & $5.07$\\
\end{tabular}}
\end{table}
Comparing DE and DE-SB, symmetry breaking consistently improves global search efficiency in all experiments. On reducable networks with a larger number of hidden neurons, as in {\bf sinc} and {\bf incsinc}, CMA-ES is superior and capable of robustly finding a solution. However, on networks with the smallest number of hidden neurons representing best compression, DE-SB outperforms CMA-ES. Furthermore, on complex problems with deeper network topologies, CMA-ES seems to have difficulties to robustly find the global optimum, even with a very large population size. On these type of problems, the true global search character of DE pays off. 
\section{Conclusions}
It is shown that symmetries in ANN-parameter space do affect the performance of Differential Evolution (DE). From theoretical considerations, we derive an ideal operator for breaking these symmetries. This ideal operator requires knowledge about the global optimum of the parameter space. Since the global optimum is not known a priori, the ideal operator is not applicable. Another concern is that a brute force implementation of the ideal operator has exponential complexity. Therefore, we propose a heuristic to approximate the ideal operator, which has negligable overhead. Unlike the CMA-ES method, which has at least quadratic complexity, the proposed DE with symmetry breaking (DE-SB) has linear complexity and is generally applicable on very high dimensional parameter spaces.

Experimental studies on a priori fixed topology networks indicate a significant improvement over standard DE in terms of required mean number of ANN-evaluations. Compared to CMA-ES, which is a state-of-the-art method, we achieve superior results especially on complex problems and smaller networks, which represent a better compression at comparable approximation quality.
We believe that other global optimization methods may also significantly benefit from symmetry breaking. Specifics of how each method should realize a symmetry breaking heuristic requires further research.

\bibliography{ann-perm-symm.bbl}
\end{document}